\documentclass[letterpaper, 10 pt, conference]{ieeeconf}  % Comment this line out if you need a4paper

\IEEEoverridecommandlockouts                              % This command is only needed if 
\usepackage{graphicx} %package to manage images
\graphicspath{ {./images/} }

\usepackage[rightcaption]{sidecap}

\usepackage{wrapfig}
\usepackage{gensymb}

\usepackage{siunitx}
\usepackage{xcolor}
\usepackage{float}

\usepackage{multirow}
\usepackage{tabu}
\usepackage{subcaption}
\usepackage{amssymb}
\usepackage{balance}

\title{\LARGE \bf
Motion Generation Interface of ROS to \\
PODO Software Framework for Wheeled Humanoid Robot                        
}

\author{
Moonyoung Lee$^{1}$, Yujin Heo$^{1}$, Saihim Cho$^{1}$\\
Hyunsub Park$^{1}$, Jun-Ho Oh$^{1}$% <-this % stops a space
\thanks{$^{1}$Is with the Humanoid Robot Research Center, Department of
Mechanical Engineering, Korea Advanced Institute of Science and Technology, 291 Daehak-ro, Yuseong-gu, Daejeon 34141, Korea 
        {\tt\small jhoh@kaist.ac.kr}}%
}

\begin{document}

\maketitle
\thispagestyle{empty}
\pagestyle{empty}

%%%%%%%%%%%%%%%%%%%%%%%%%%%%%%%%%%%%%%%%%%%%%%%%%%%%%%%%%%%%%%%%%%%%%%%%%%%%%%%%
\begin{abstract}

This paper discusses the development of robot motion generation interface between a real-time software architecture and a non-real-time robot operating system. 
In order for robots to execute intelligent manipulation or navigation, close integration of high-level perception and low-level control is required. However, many available open-source perception modules are developed in ROS, which operates on Linux OS that don't guarantee RT performance. This can lead to non-deterministic responses and stability problems that can adversely affect robot control. As a result, many robotic systems devote RTOS for low-level motion control. Similarly, the humanoid robot platform developed at KAIST, Hubo, utilizes a custom real-time software framework called PODO. Although PODO provides easy interface for motion generation, it lacks interface to high-level frameworks such as ROS. 
As such, we present a new motion generation interface between ROS and PODO that enables users to generate motion trajectories through standard ROS messages while leveraging a real-time motion controller.
With the proposed communication interface, we demonstrate series of manipulator tasks on the actual  wheeled humanoid platform, M-Hubo. The overall communication interface responsiveness was at most 27 milliseconds.
\end{abstract}
%%%%%%%%%%%%%%%%%%%%%%%%%%%%%%%%%%%%%%%%%%%%%%%%%%%%%%%%%%%%%%%%%%%%%%%%%%%%%%%%
\section{INTRODUCTION}

In the past couple decades, with the adoption of industrial robots for assembly and manufacturing applications, the field of precise manipulator control has become well-established. 
In addition to precise control, customer demands for robotic applications have recently grown to entail significant degree of autonomy and intelligence. This trend has been vigorously driven by reduced sensor costs, enhanced GPU performances, and reduced development time due to available open-source machine-learning tools.
While there are large efforts to meet these demands, majority of these works are done with a fixed-based 6 D.O.F. manipulator arm, commonly phrased as a collaborative robot. On a humanoid platform, in which the robot-base is not stationary and manipulation environment is highly dynamic, integrating precise control with high degree of autonomy becomes challenging.

\begin{figure}[h]
    \centering
    \includegraphics[scale=0.5]{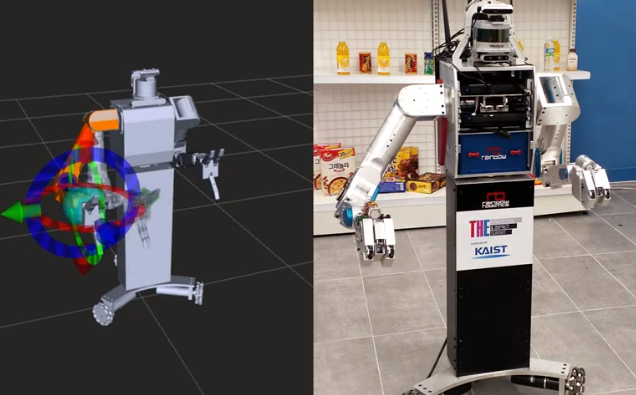}
    \caption{Motion Generation through ROS API for the wheeled humanoid M-Hubo}
    \label{fig:moveit_robot}
\end{figure}

Of many, one primary challenge of developing a complex autonomous system is the robot's software framework. The software framework must be capable of interfacing with the low-level hardware and executing precise control tasks. In addition, the software framework must also be capable of handling high-level planning tasks that are computation-heavy from processing dense visualization data. Precise control requires for a framework with guaranteed real-time (RT) performance that can handle low-level sensor and motor controller interface~\cite{xbotcore}. Typically, such frameworks handle deadline constraints at high frequencies close to 1 kHz. 

On the other hand, high-level autonomous tasks are computationally demanding and operate at lower frequencies close to 10Hz. Naturally, tools built for autonomy related tasks are fundamentally different and exist in non-real-time (NRT) framework. As such, in order to achieve precise control with high degree of autonomy, the problem of integrating NRT and RT frameworks is introduced.

% Work Goal
To address this issue, we developed an API that provides a ROS interface for generating motion for the wheeled humanoid M-Hubo platform developed in KAIST Hubo Lab.
This interface abstracts the low-level motion control implemented in the custom real-time PODO software, such as trajectory and IK solvers, and hence allows developers to focus on high-level task planning for more complex sequence of behaviors.
We discuss in depth the implementation, evaluation, and challenges of creating a heterogeneous software to interface between NRT and RT frameworks. 

% Work Summary
We demonstrate our proposed framework on the actual robot by measuring the communication delay and the accuracy of the robot's motion from the position, velocity, and acceleration profile.

Our main contributions are:
\begin{itemize}
    \item Seamless API interface of ROS middleware to custom RT PODO framework. This is a significant contribution to our previous work~\cite{Mhubo} by creating a new method of interface for motion generation.
    \item Evaluation and demonstration of the proposed interface on a wheeled humanoid platform M-Hubo.
    
\end{itemize}
% The remaining of the paper is structured as follows: 
% We review previous works and compare them briefly to our system in section II.
% Section III summarizes the robot hardware platform and software architecture developed for handling high-level tasks with real-time motion controllers. 
% Section IV discusses the software interfaces and communication handling.
% We conclude by discussing experiment and results from demonstrating the proposed framework on the actual robot in Section V. 

%%%%%%%%%%%%%%%%%%%%%%%%%%%%%%%%%%%%%%%%%%%%%%%%%%%%%%%%%%%%%%%%%%%%%%%%%%%%%%%%
\section{RELATED WORK}

One of the widely adopted types of robot control system is an open-source framework without RT performance. Due to the large scope of interdisciplinary fields involved in creating a robotic system, there has been significant efforts to create a modular and widely-adopted robot framework such as ROS, YARP, OROCOS~\cite{ros}~\cite{yarp}. 

These are developed on the popular Linux OS that don't guarantee RT performance, which can result to non-deterministic responses and stability problems such as priority inversion, which can adversely affect robot control.
An example work that utilize this system is a small-sized humanoid NimboRo-Op platform~\cite{nimboro}. This work utilizes ROS-based framework on standard Linux kernel with high-precision timerfd API that enhances performance of meeting timing constraints to control the robot at maximum of 125Hz rate. 
Similarly, Fetch \& Freight~\cite{fetch}, the newer mobile manipulator platform from the developers of the popular PR2, does not meet RT guarantees by using RS-485 communication protocol. For bigger scale humanoids, in which precision becomes more critical, RT guarantee is required especially for high-frequency control periods. As such, there has been on-going research for development of ROS 2.0, which integrates the traditional ROS framework with RT capabilities. However, as evaluated in this work ~\cite{ros2}, although the current beta version of ROS 2.0 is able to achieve soft and firm RT, it is not able to meet hard RT due to traffic limitations in the Linux Network Stack.

This leads us to discuss frameworks that utilize real-time operating system (RTOS) such as Windows CE, RTAI, and Xenomai. In contrast to the standard Linux kernel which schedules based on fairness, RTOS schedules based on priorities, where lower priority tasks can be pre-empted. This ensures deterministic handling of control-related tasks with predefined time.   
XBotCore~\cite{xbotcore} is a RT software framework that also provides API in the format of plugin-handlers for interacting with the NRT middleware. Although this work provides a hard RT control system with external framework integration for NRT middleware, it is not suited for our application because XBotCore is limited to hardware systems that utilize EtherCAT communication.

A similar heterogeneous framework to ours is proposed in~\cite{snu}. A RT architecture based on ROS and Xenomai is developed by adapting a communication interface of the cross-domain datagram protocol (XDDP). This system requires modifying device drivers, which for the proposed simple embedded Raspberry-Pi system is feasible, but for full-scale humanoid systems with multiple PCs can become burdensome.
Our previous work of PODO~\cite{Mhubo} also operates as a heterogeneous framework but only receives minimal visual data from ROS. Not only was the previous framework limited to one-way communication but it did not support any motion generation requests. As such, we propose our current work that provides fully expanded API.

%%%%%%%%%%%%%%%%%%%%%%%%%%%%%%%%%%%%%%%%%%%%%%%%%%%%%%%%%%%%%%%%%%%%%%%%%%%%%%%%
\section{SYSTEM SETUP}

\subsection{Robot Platform Hardware }

\begin{figure}[h]
    \centering
    \includegraphics[scale=0.8]{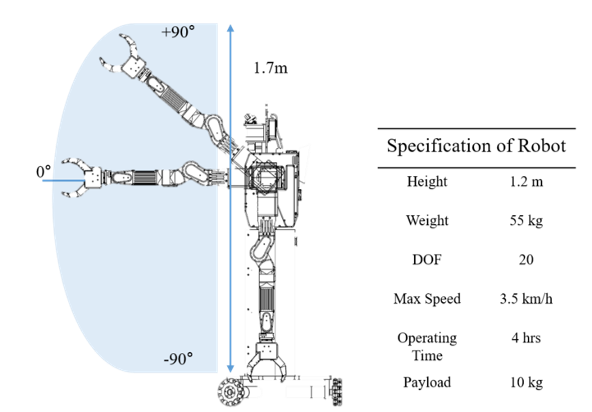}
    \caption{Workspace and overall robot specification}
    \label{fig:workspace}
\end{figure}

The wheeled humanoid M-Hubo consists of 20 degrees of freedom total. It contains two 7 DOF manipulator arms, which have optimized joint limits to provide maximum workspace of $100$\degree each with $0.8$ m reach. The authors utilize the manipulator design from the prior DRC-HUBO+ humanoid robot~\cite{jungDRC}, which enables precise position control due to high rigidity in mechanical design and minimal jitter in communication of robot joint references. 
The omni-directional base, which can reach a maximum speed of 3.5 km/h, provides effective locomotion control for variety of automated manipulation tasks. 

\subsection{Software Architecture }
The software architecture is intuitively divided among perception and motion control. For subtasks related to high-level perception and task planning, we dedicate a separate Vision PC operating on ROS Kinetic. For RT motion control, we dedicate a separate Motion PC operating with PODO ~\cite{podo}. 
\begin{figure}[h]
    \centering
    \includegraphics[scale=0.4]{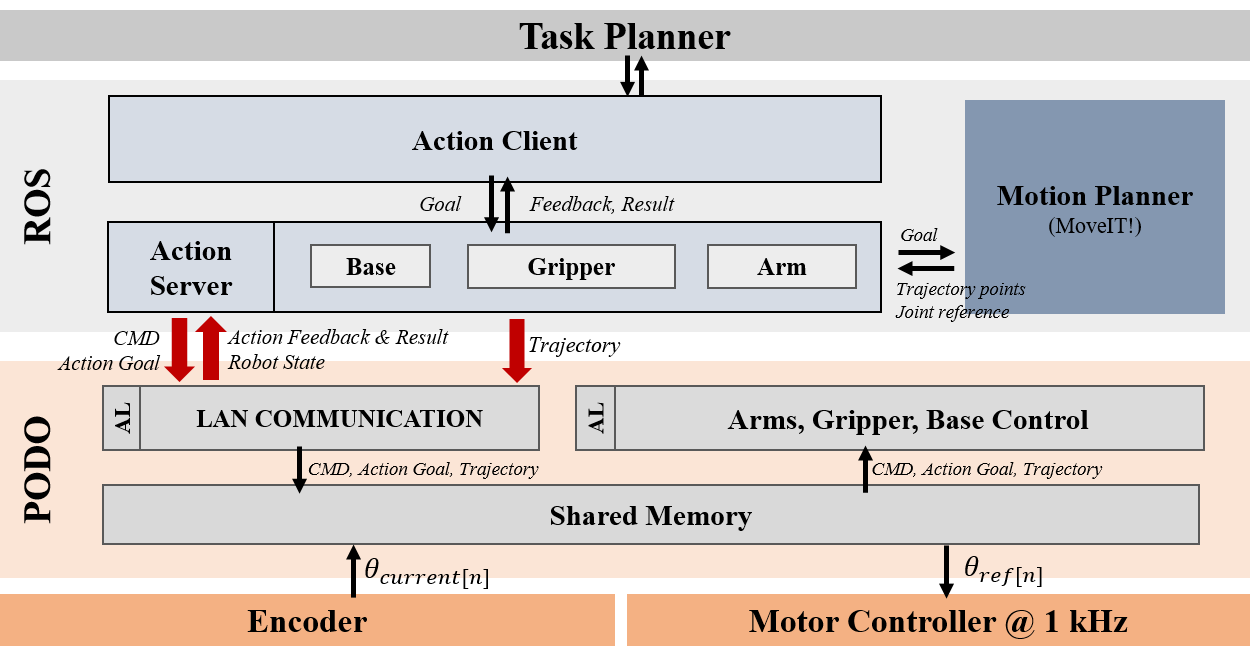}
    \caption{Overall framework for interfacing ROS for high-level tasks and PODO for low-level motion control}
    \label{fig:software}
\end{figure}

\begin{figure*}[t!]
    \centering
    \includegraphics[scale=0.6]{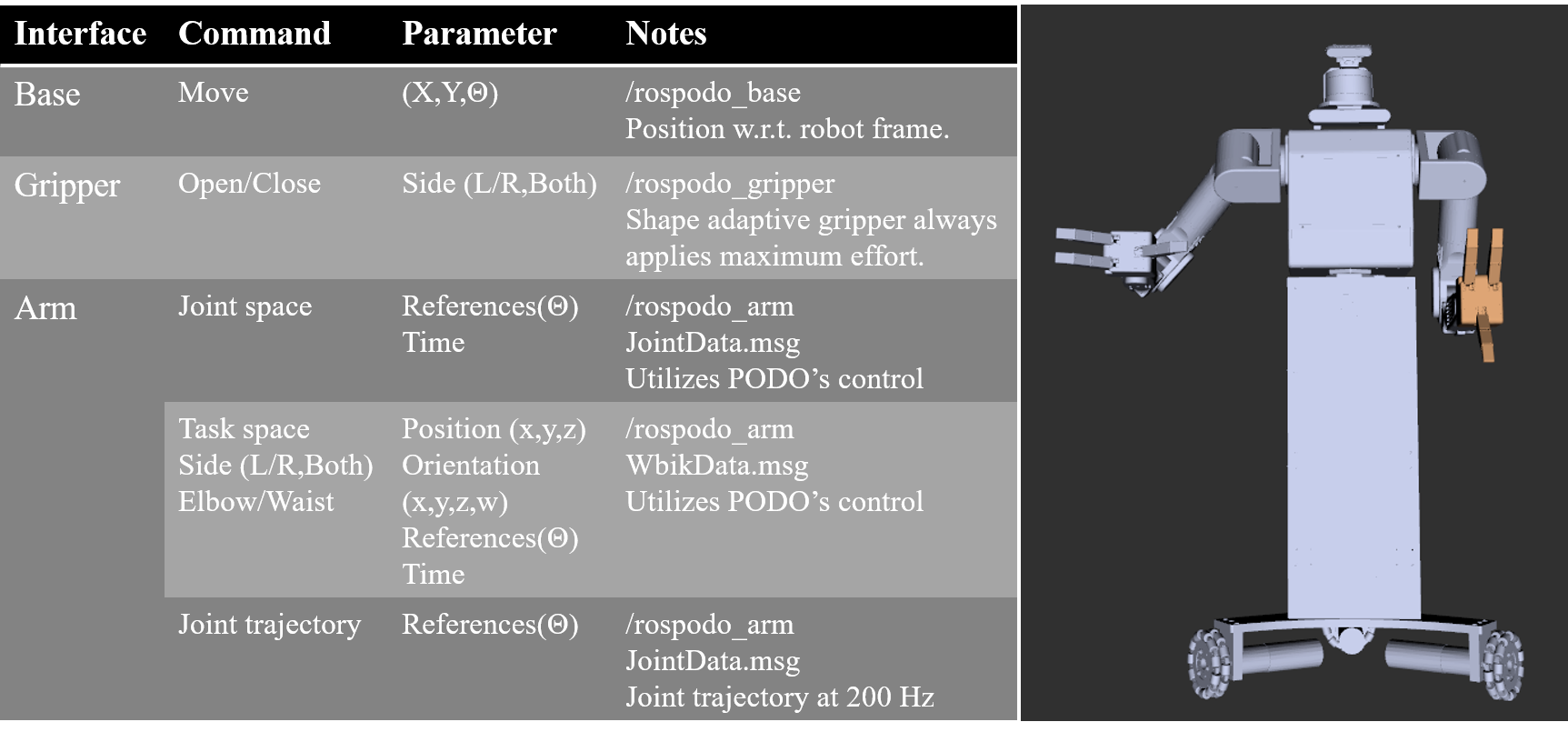}
    \caption{API interface for the M-Hubo platform. Three separate threads are dedicated for Base, Gripper, and Arm, allowing for concurrent utilization of multiple parts of the robot. }
    \label{fig:API}
\end{figure*}

Within the Vision PC exist the ROS Action Server and Client responsible for handling motion requests through the API interface. The API receives inputs from a high-level motion planner in ROS and transmits them to the Motion PC through TCP/IP socket write. 

The Motion PC translates the high level requests into manipulator or base trajectories through the motion controller operating at 200Hz period. These are then additionally handled by the low-level motor controllers to create joint references operating at 1kHz period. The PODO software reduces communication jitter by ensuring accuracy of joint information and sensor updates by synchronizing multiple threads at very regular intervals through a CAN bus interface. The RT Linux interface is based on Xenomai on Ubuntu 16.04 (Xenial) running on Intel NUK6i7KYK PC with i7-6770 4-core processor at 3.5 GHz and 8GB RAM. The overall software framework is depicted in Figure~\ref{fig:software}.

%%%%%%%%%%%%%%%%%%%%%%%%%%%%%%%%%%%%%%%%%%%%%%%%%%%%%%%%%%%%%%%%%%%%%%%%%%%%%%%%
\section{Software Interface and Communication}
The following sections below discuss the communication implementations within ROS, between ROS and PODO, and within PODO. 

\subsection{ROS API for M-Hubo}
The API utilizes standard ROS actionlib package to control robot's base, arms, and grippers. These three entities can be called concurrently, allowing for complex motions. When selecting the interface for each entity, the user can provide command, data, and additional parameters such as left/right arm motions.
Although this API already abstracts the low-level control implemented in the PODO software, such as trajectory and IK solvers, it can also utilize third party tools in ROS to directly generate joint-space trajectory for the arms. 
 The interfaces for base, grippers, and arms are shown in Figure~\ref{fig:API}. 

%%%%%%%%%%%%%%%%%%%%%%%%%%%%%%%%%%%%%%%%%%%%%%%%%%%%%%%%%%%%%%%%%%%%%%%%%%%%%%%%
\subsection{Communication Handling in ROS}

With the proposed framework, users can easily request motions from the API by creating an action client appropriate for their application. The action client evokes function callback with the desired parameters from the action server.
To handle base, gripper, and arm requests, the action server contains three independent threads running at 200Hz. While the base and gripper have simple single-request goals, the interface for the arms is more complex to handle both single-request and high-frequency requests.

\begin{figure}[h]
    \centering
    \includegraphics[scale=0.6]{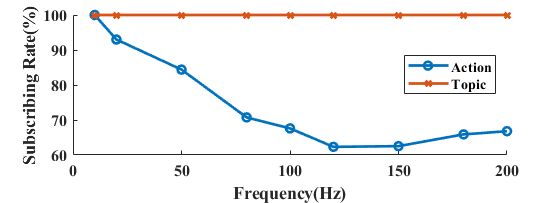}
    \caption{Packet transmission success rate using Action messages and topics containing joint references}
    \label{fig:action_joint_states_plot}
\end{figure}

Although single-request function calls resulted in negligible communication delays between frameworks, resulting in a smooth motion on the actual robot, high-frequency-request function calls resulted in communication delays, resulting in a non-smooth motion on the actual robot. Communication delay between NRT \& RT software frameworks are discussed in section below.    

High-frequency-requests of manipulating the arms from the action client to the server were initially designed by wrapping the standard /joint\_states topics as action messages to utilize the goal, feedback, and result interface. The /joint\_states topics contain position, velocity, and effort data of robot joints.

However, directly wrapping and passing the /joint\_states topics published by MoveIt at 10Hz as desired reference to the robot, which operates at 200Hz control periods, would result in uneven motion trajectory due to  many discontinuities in both position and velocity profiles. As such, we added a simple interpolation in the client to match the communication periods. 
It is interesting to note that we observed significant drop in successful packet-transmission-rate between client and server at high-frequencies above 200Hz, which is the control period for our robot. As shown in Figure~\ref{fig:action_joint_states_plot}, 33\% of the packet transmitted result in unsuccessful action responses between client and server at 200Hz. In contrast, ROS /joint\_states topics resulted in 100\% subscription rate at even higher frequencies. However, to minimize the effect of communication delay, we forego the high-frequency request for enabling dynamic control in ROS, and implement a single-request trajectory array containing joint states references over time.

%%%%%%%%%%%%%%%%%%%%%%%%%%%%%%%%%%%%%%%%%%%%%%%%%%%%%%%%%%%%%%%%%%%%%%%%%%%%%%%%
\subsection{Communication Handling between Frameworks}
Ideal communication between frameworks should have immediate response time with constant repeatability. A homogenous real-time framework can guarantee repeatable communication. The communication response time can then be determined through hardware selection such as EtherCAT, RS-485, or CAN. For a heterogeneous framework such as the one proposed in this paper, although a TCP/IP socket write is requested at 200Hz interval from ROS server, thread execution priority handled by the OS only approximately meets the requested deadline. 

\begin{figure}[h]
    \centering
    \includegraphics[scale=0.55]{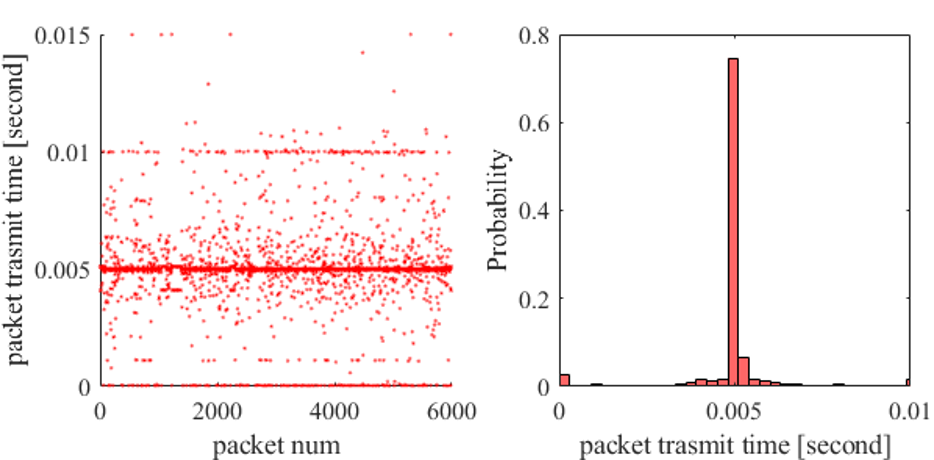}
    \caption{Distribution of packet transmit time between frameworks}
    \label{fig:packet transmit time distribution}
\end{figure}

Upon quick inspection of communication time using Wireshark, the authors could verify most packet times were under 5 milliseconds. However, closer inspection with a time distribution of packet requests within ROS framework, as shown in Figure~\ref{fig:packet transmit time distribution}, raises an issue of communication delay as expected across frameworks. The histogram illustrates that more than 10\% of the packet sent have transmit time greater than 5 milliseconds. We also observe small clumping of packets sent in either 1 tick previous or after the desired period.

\begin{figure}[h]
    \centering
    \includegraphics[scale=0.6]{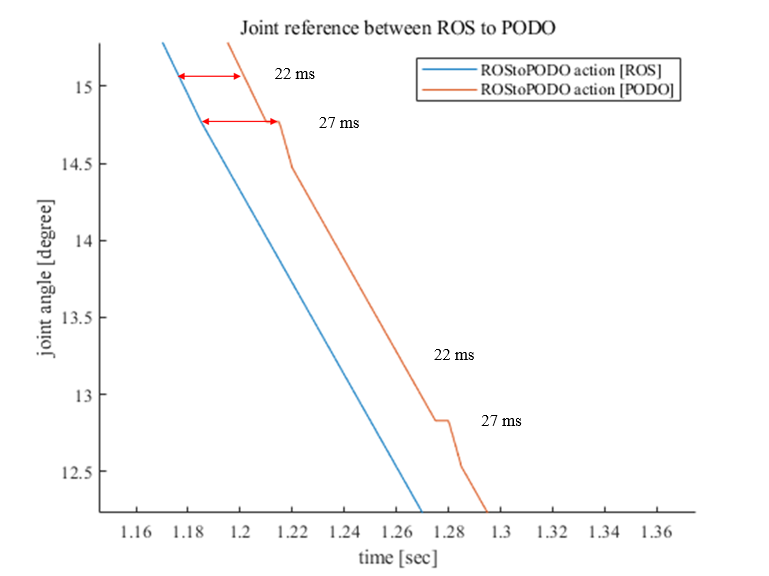}
    \caption{Desired robot state joint reference from ROS to PODO. Due to NRT communication, differences in packet transmission time can be observed.}
    \label{fig:ros2podo_delay}
\end{figure}

 For a real-time framework such as PODO, these deviation in packet transmission times due to the nature of soft deadlines result in corresponding request to be handled in the next time tick in respect to PODO. Because PODO framework directly utilizes the given joint references as inputs to the motor controller which operates at 1 kHz, the difference of 1 tick in communication adversely affects the low level control. As show in Figure~\ref{fig:ros2podo_delay}, majority of transmission time between ROS and PODO is 22 milliseconds while packets that are delayed and handled in the next tick require 27 milliseconds.

\begin{figure}[h]
    \centering
    \includegraphics[scale=0.6]{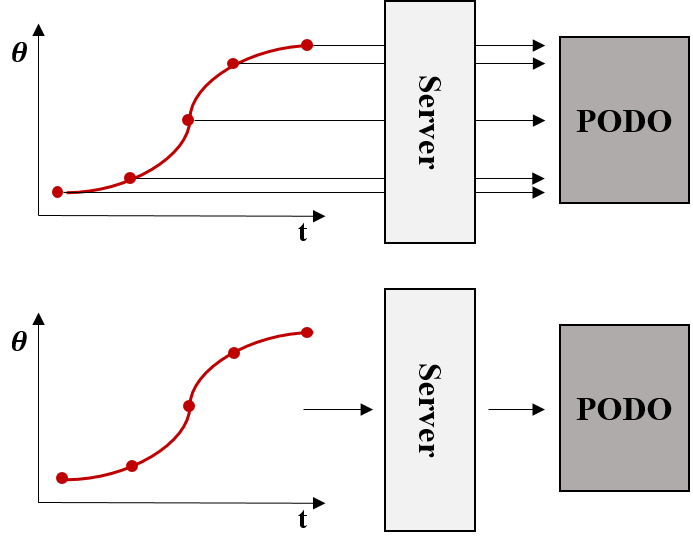}
    \caption{High-frequency-request interface and a single-request from ROS to PODO framework}
    \label{fig:trajectorypicture}
\end{figure}

Without significantly modifying the overall heterogeneous software framework, this issue of non-guaranteed real-time deadlines cannot be resolved. As a result, as mentioned in the section above, the authors limited the extent of the API capability. Instead of handling high-frequency requests for enabling dynamic control in ROS, the action client receives a complete trajectory array over time and instead issues a single-request to PODO as shown in Figure~\ref{fig:trajectorypicture}.
This resolves the communication delay of utilizing a heterogeneous framework for the cost of limiting the use case. However this is not a critical compromise since most motion planners that generate motion trajectories require couple seconds of static planning.  

%%%%%%%%%%%%%%%%%%%%%%%%%%%%%%%%%%%%%%%%%%%%%%%%%%%%%%%%%%%%%%%%%%%%%%%%%%%%%%%%
\subsection{Communication Handling in PODO}
Once ROS server issues a single-request trajectory through TCP/IP socket write, the PODO framework is responsible for handling that reference so that the low-level motor controller can generate precise robot motion.

\begin{figure}[h]
    \centering
    \includegraphics[scale=0.45]{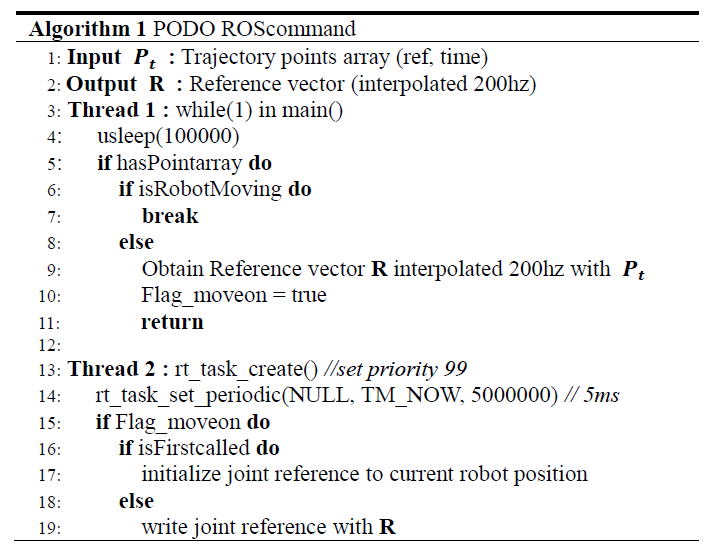}
    \label{fig:algorithm}
\end{figure}

PODO receives the data through a separate thread called PODOLAN for handling communication between frameworks. 
The received data is then made accessible to the multiple processes within PODO framework through the Shared Memory. 
Of the many processes, or AL, that can generate robot motion for specific tasks, we developed a dedicated AL that generates motion based on references passed from ROS API.  

\begin{figure}[h]
    \centering
    \includegraphics[scale=0.65]{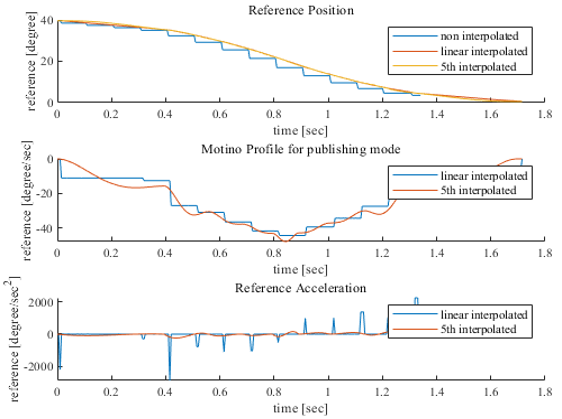}
    \caption{Effect of interpolation on the motion profile of the received trajectory}
    \label{fig:motionprofile}
\end{figure}

This AL's purpose is to interpolate the requested trajectory to match robot control frequency of 200Hz.
To do so, we utilize a 5th-order polynomial interpolation of joint references over time to generate velocity and acceleration profiles without discontinuities as shown in Figure~\ref{fig:motionprofile}.

%%%%%%%%%%%%%%%%%%%%%%%%%%%%%%%%%%%%%%%%%%%%%%%%%%%%%%%%%%%%%%%%%%%%%%%%%%%%%%%%
\section{EXPERIMENT AND RESULTS}

\begin{figure*}[t]
    \centering
    \includegraphics[scale=0.45]{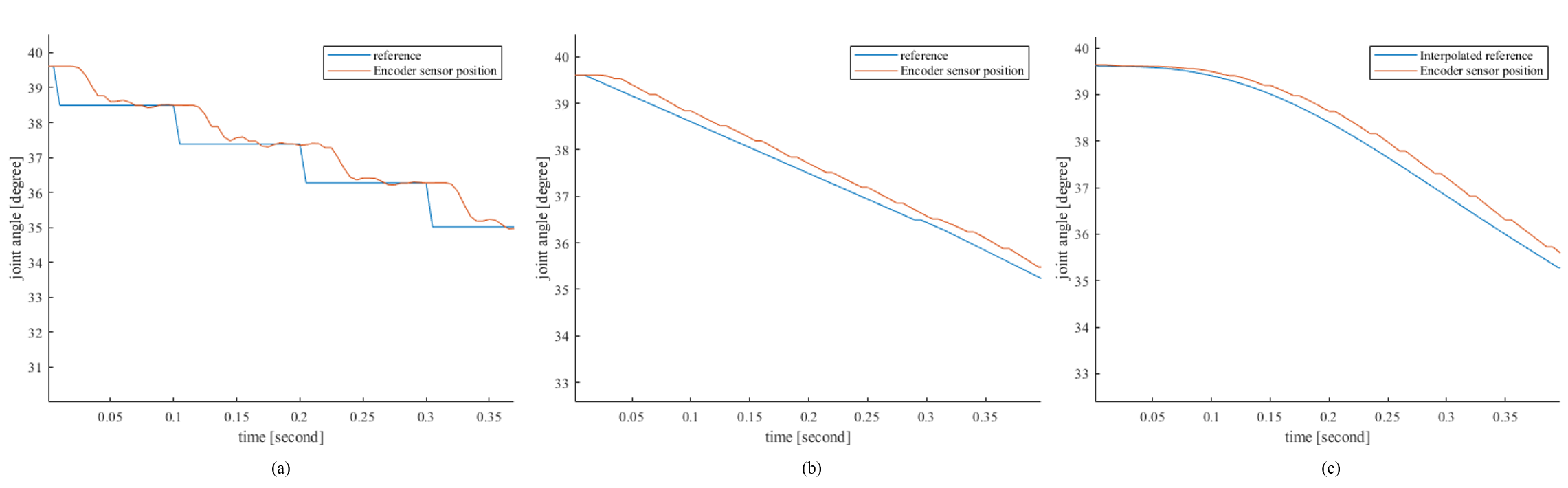}
    \caption{Plot of requested references and measured joint values across three separate communication interface. (a) low-frequency requests of 10Hz, (b) high-frequency requests of 200Hz, (c) single-request trajectory}
    \label{fig:encoder_plot}
\end{figure*}

To validate and evaluate the performance of our API for a heterogeneous framework, we perform a set of experiments by generating motions in ROS to operate the M-hubo platform.  
The overall performance of the ROS interface with PODO is evaluated by the responsiveness and accuracy of the motion trajectory generated over the real robot.

\begin{figure}[h]
    \centering
    \includegraphics[scale=0.6]{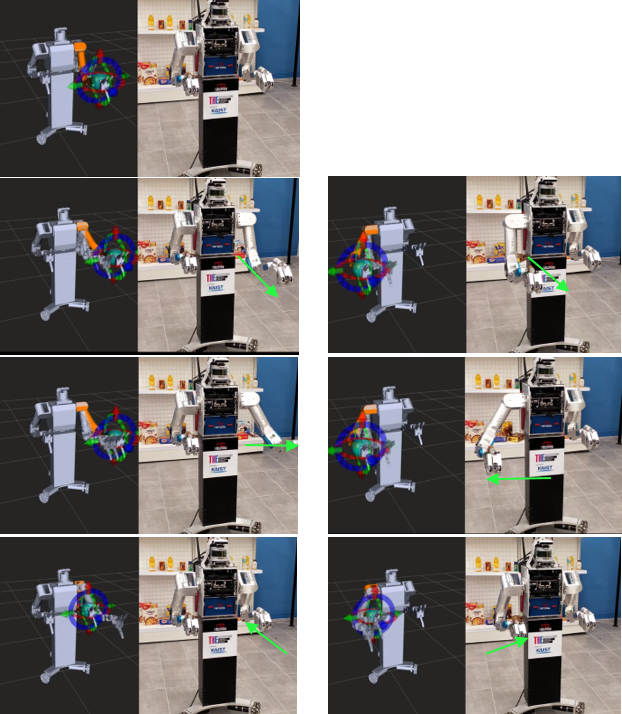}
    \caption{Timelapse depicting the robot following the requested motion trajectories generated in ROS}
    \label{fig:timelapse}
\end{figure}

We utilize the MoveIt! package with the OMPL library~\cite{moveit} to generate a series of collision-free trajectories for the robot’s manipulators. These requested trajectories generated in ROS are then handled through our proposed API to evoke the motion controller in PODO for actual robot movement. Figure~\ref{fig:timelapse} illustrates the robot following the requested trajectories in a time-lapse manner.

To summarize the communication responsiveness from section IV, the bottleneck of communication delay occurs during ROS and PODO communication. While there is negligible delay in communication within frameworks, between NRT \& RT framework communication can be up to 27 milliseconds. 
To evaluate the accuracy of the motion executed, we measure the error, defined as the difference between the desired joint reference requested from ROS and the actual joint references measured from the robot encoders. For the same requested trajectory, we compare three separate communication handling: low-frequency requests of 10Hz, high-frequency requests of 200Hz with interpolation in ROS, and single-request trajectory with interpolation in PODO. The differences between the requested joint references and the actual robot joint encoder values are depicted in Figure~\ref{fig:encoder_plot}.

The error across the three communication handling, measured in standard deviation, are 1.520\degree, 0.039\degree, and 0.061\degree respectively. 
As expected, the error between requested reference and actual measured joints was more than 20x higher at low-frequency requests than high-frequency requests due to the discontinuities in the motion profile. 
What was unexpected was that the single-request trajectory interface performed no better than the previous high-frequency request interface, which contain communication delay mentioned above. 
The authors infer that this error is dominantly affected by the control algorithm utilized rather than the communication delay of the API. For further reducing this error, we can fine-tune the parameters for motion control, which is outside the scope of this paper.

%%%%%%%%%%%%%%%%%%%%%%%%%%%%%%%%%%%%%%%%%%%%%%%%%%%%%%%%%%%%%%%%%%%%%%%%%%%%%%%%
\section{CONCLUSIONS}

In this paper we presented a motion generation interface of ROS middleware to a real-time software framework. 
This is achieved through a seamless API integration of NRT ROS to our custom RT software PODO. 
With the proposed API interface, users can generate motion trajectories for the actual wheeled humanoid platform M-Hubo through standard ROS messages and leverage and real-time motion controller functionality of PODO.
We implement various communication interfaces for this heterogeneous software framework, and evaluate them based on the responsiveness and accuracy of the motion trajectory generated over the real robot.

With the proposed communication interface, we demonstrated successful implementation by executing series of manipulator tasks on the actual robot from trajectory requests generated from ROS.
The communication interface responsiveness was approximately 27 milliseconds and the accuracy error, defined as the standard deviation between the desired joint reference and measured robot's joint values, was 0.06\degree. 
The proposed software framework is not robot-independent. For future works, the framework will be made modular and be expanded to be robot agnostic to provide API for variety of robot platforms.

The proposed API for interfacing ROS and PODO frameworks for the wheeled humanoid platform M-hubo is made available as open-source. All related manual, code, and video can be found on the github page \textit{github.com/KaistInstitute/ros\_podo\_connector}

%%%%%%%%%%%%%%%%%%%%%%%%%%%%%%%%%%%%%%%%%%%%%%%%%%%%%%%%%%%%%%%%%%%%%%%%%%%%%%%%
\section{ACKNOWLEDGEMENT}
This work was supported by the Technology Innovation Program (or Industrial Strategic Technology Development Program (20001881, Development of Robotic Grasping, manipulation and control technology for human tools based on multi-modal recognition) funded By the Ministry of Trade, industry \& Energy(MOTIE, Korea).

\addtolength{\textheight}{-12cm}   % This command serves to balance the column lengths

%%%%%%%%%%%%%%%%%%%%%%%%%%%%%%%%%%%%%%%%%%%%%%%%%%%%%%%%%%%%%%%%%%%%%%%%%%%%%%%%

%%%%%%%%%%%%%%%%%%%%%%%%%%%%%%%%%%%%%%%%%%%%%%%%%%%%%%%%%%%%%%%%%%%%%%%%%%%%%%%%

%%%%%%%%%%%%%%%%%%%%%%%%%%%%%%%%%%%%%%%%%%%%%%%%%%%%%%%%%%%%%%%%%%%%%%%%%%%%%%%%

% \section*{ACKNOWLEDGMENT}

% This work was supported by Development of Core Technologies and a Standard Platform for Humanoid Robot [10060103], project from the Ministry of Trade, Industry and Energy (MOTIE) of the Republic of Korea 

% \begin{thebibliography}{99}

% \bibitem{c1} G. O. Young, �Synthetic structure of industrial plastics (Book style with paper title and editor),� 	in Plastics, 2nd ed. vol. 3, J. Peters, Ed.  New York: McGraw-Hill, 1964, pp. 15�64.

% \end{thebibliography}

\bibliographystyle{./IEEEtran} % use IEEEtran.bst style
\bibliography{./egbib2}

\end{document}